\documentclass[11pt,a4paper]{article}
\usepackage{naaclhlt2019}

\usepackage[T1]{fontenc}
\usepackage[utf8]{inputenc}
\usepackage{graphicx}
\usepackage{ragged2e}  %
\usepackage{placeins}
\usepackage{tabu}
\usepackage{url}
\usepackage{latexsym}
\usepackage{booktabs}
\usepackage{amsmath,amsfonts}
\usepackage{braket}
\usepackage{multirow}
\usepackage{titlesec}
\usepackage{relsize}
\usepackage{colortbl}
\usepackage{lipsum}
\usepackage{fancyhdr}
\usepackage{enumitem}

\newcommand{\tbf}[1]{\textbf{#1}}

\aclfinalcopy %
\title{Improving the Coverage and the Generalization Ability of \\ Neural Word Sense Disambiguation through Hypernymy and Hyponymy Relationships}

\author{Loïc Vial \quad Benjamin Lecouteux \quad Didier Schwab}

\date{}

\begin{document}

\maketitle

\begin{abstract}
In Word Sense Disambiguation (WSD), the predominant approach generally involves a supervised system trained on sense annotated corpora. The limited quantity of such corpora however restricts the coverage and the performance of these systems. In this article, we propose a new method that solves these issues by taking advantage of the knowledge present in WordNet, and especially the hypernymy and hyponymy relationships between synsets, in order to reduce the number of different sense tags that are necessary to disambiguate all words of the lexical database. Our method leads to state of the art results on most WSD evaluation tasks, while improving the coverage of supervised systems, reducing the training time and the size of the models, without additional training data. In addition, we exhibit results that significantly outperform the state of the art when our method is combined with an ensembling technique and the addition of the WordNet Gloss Tagged as training corpus.
\end{abstract}

\section{Introduction}

Word Sense Disambiguation (WSD) is a task which aims to clarify a text by assigning to each of its words the most suitable sense labels, given a predefined sense inventory.

Various approaches have been proposed to achieve WSD, and they are generally ordered by the type and the quantity of resources they use:
\begin{itemize}
    \item \tbf{Knowledge-based} methods rely on dictionaries, lexical databases, thesauri or knowledge graphs as main resources, and use algorithms such as lexical similarity measures \citep{Lesk1986} or graph-based measures \citep{Moro2014EntityLM}.
    \item \tbf{Supervised} methods, on the other hand, exploit sense annotated corpora as training instances that can be used by a multi-class classifier such as SVM \citep{Chan2007,Zhong2010}, or more recently by a neural network \citep{kaageback2016word}.
    \item \tbf{Semi-supervised} methods generally use unannotated data to artificially increase the quantity of sense annotated data and hence improve supervised methods \citep{yuan_2016}.
\end{itemize}

Supervised methods are by far the most predominant as they generally offer the best results in evaluation campaigns (for instance \citep{Navigli2007}). State of the art classifiers used to combine a set of specific features such as the parts of speech tags of surrounding words, local collocations \citep{Zhong2010} and pretrained word embeddings \citep{iacobacci2016embeddings}, but they are now replaced by recurrent neural networks which learn their own representation of words \citep{raganato2017,minh2018}.

One of the major bottleneck of supervised systems is the restricted quantity of manually sense annotated corpora. Indeed, while the lexical database WordNet \citep{miller1995wordnet}, the sense inventory of reference used in most works on WSD, contains more than 200\,000 different word-sense pairs\footnote{\url{https://wordnet.princeton.edu/documentation/wnstats7wn}}, the SemCor \citep{Miller1993}, the corpus which is used the most in the training of supervised systems, only represents approximately 34\,000 of them.

Many works try to leverage this problem by creating new sense annotated corpora, either automatically \citep{pasini2017}, semi-automatically \citep{taghipourng2015}, or through crowdsourcing \citep{yuan_2016}, but in this work, the idea is to solve this issue by taking advantage of one of the multiple semantic relationships between senses included in WordNet: the hypernymy and hyponymy relationships. Our method is based on three observations:
\begin{enumerate}
    \item A sense, its hypernym and its hyponyms share a common idea or concept, but on different levels of abstraction.
    \item In general, a word can be disambiguated using the hypernyms of its senses, and not necessarily the senses themselves. 
    \item Consequently, we do not need to know every sense of WordNet to disambiguate all words of WordNet. 
\end{enumerate} 
\tbf{Contributions:}
We propose a method for reducing the vocabulary of senses of WordNet by selecting the minimal set of senses required for differentiating the meaning of every word. 
By using this technique, and converting the sense tags present in sense annotated corpora to the most generalized sense possible, we are able to greatly improve the coverage and the generalization ability of supervised systems.

We start by presenting the state of the art of supervised neural architectures for word sense disambiguation, then we describe our new method for sense vocabulary reduction. Our method is then evaluated by measuring its contribution to a state of the art neural WSD system evaluated on classic WSD evaluation campaigns.

The code for using our system or reproducing our results is available at the following URL:\newline
\url{https://github.com/getalp/disambiguate}

\section{Neural Word Sense Disambiguation}

The neural approaches for WSD fall into two categories: approaches based on a neural model that learns to classify a sense directly, and approaches based on a neural language model that learns to predict a word, and is then used to find the closest sense to the predicted word.

\subsection{Language Model Based WSD}

The core of these approaches is a powerful neural language model able to predict a word with consideration for the words surrounding it, thanks to a recurrent neural network trained on a massive quantity of unannotated data. The main works that implement these kind of model are \citet{yuan_2016} and \citet{minh2018}.

Once the language model is trained, its predictions are used to produce sense vectors as the average of the word vectors predicted by the language model in the places where the words are sense annotated.

At test time, the language model is used to predict a vector according to the surrounding context, and the sense closest to the predicted vector is assigned to each word.

These systems have the advantage of bypassing the problem of the lack of sense annotated data by concentrating the power of abstraction offered by recurrent neural networks on a good quality language model trained in an unsupervised manner.
However, sense annotated corpora are still indispensable to construct the sense vectors, and the quantity of data needed for training the language model (100 billion tokens for \citet{yuan_2016}, 2 billion tokens for \citet{minh2018}) makes these systems more difficult to train than those relying on sense annotated data only. 

\subsection{Classification Based WSD}\label{sec:classif_based_wsd}

In these systems, the main neural network directly classifies and attributes a sense to each input word. Sense annotations are simply seen as tags put on every word, like a POS-tagging task for instance.

These models are more similar to classical supervised models such as \citet{Chan2007}, except that the input features are not manually selected, but trained as part of the neural network (using pre-trained word embeddings or not).

In addition, we can distinguish two separate branches of these types of neural networks: 
\begin{enumerate}[leftmargin=*]
    \item Those in which we have several distinct and small neural networks (or classifiers) for every different word in the dictionary \citep{iacobacci2016embeddings,kaageback2016word}, each of them being able to manage a particular word and its particular senses. For instance, one of the classifiers is specialized into choosing between the four possible senses of the noun ``mouse''. This type of approaches is particularly fitted for the lexical sample tasks, where a small and finite set of very ambiguous words have to be sense annotated in several contexts, but it can also be used in all-words word sense disambiguation tasks.
    \item Those in which we have a bigger and unique neural network that is able to manage all different words and assign a sense in the set of all existing sense in the dictionary used \citep{raganato2017}.
\end{enumerate}

The advantage of the first branch of approaches is that in order to disambiguate a word, limiting our choice to one of its possible senses is computationally much easier than searching through all the senses of all words. To put things in perspective, the number of senses of each word in WordNet ranges from 1 to 59,
whereas the total number of senses considering all words is 206\,941. 

The other approach however has an interesting property: all senses reside in the same vector space and hence share features in the hidden layers of the network. This allows the model to predict a common sense for two different words (i.e. synonyms), but it also offers the possibility to predict a sense for a word not present in the dictionary (e.g. neologism, spelling mistake...), and let the user or the underlying system to decide afterwards what to do with this prediction.

In practice, this ability of merging multiple sense tags together is especially useful when working with WordNet: indeed, this lexical database is based on the notion of synonym sets or ``synsets'', group of senses with the same meaning and definition. Disambiguating with synset tags instead of sense tags is a common practice \citep{yuan_2016,minh2018}, as it effectively decreases the output vocabulary of the classifier that considers all senses in WordNet from 206\,941 to 117\,659, and one can unambiguously retrieve the sense tag given a synset tag and the tagged word (because every sense of a word belong to a different synset).

In this work, we go further into this direction and we present a method based on the hypernymy and hyponymy relationships present in WordNet, in order to merge synset tags together and reduce even more the output vocabulary of such neural WSD systems.

\section{Sense Vocabulary Reduction} 

We can draw three issues of the current situation regarding supervised WSD systems:

\begin{enumerate}
    \item The training of systems that directly predict a tag in the set of all WordNet senses becomes slower and take more memory the larger the output vocabulary is. This output layer size going up to 206\,941 if we consider all word-senses, and 117\,659 if we consider all synsets.%
    \item Due to the small number of manually sense annotated corpora available, a target word may never be observed during the training, and therefore the system would not be able to annotate it.%
    \item For the same reason, a word may have been observed, but not all of its senses. In this case the system is able to annotate the word, but if the expected sense has never been observed, the output will be wrong, regardless of the architecture of the supervised system.
\end{enumerate}

In the SemCor \citep{Miller1993} for instance, the largest manually sense annotated corpus available, 
words are annotated with 33\,760 different sense keys, which corresponds to approximately 16\% of the sense inventory of WordNet\footnote{\url{https://wordnet.princeton.edu/documentation/wnstats7wn}}.

Grouping together multiple senses is hence a good way to overcome all these issues: by considering that multiple tags refer in fact to the same concept, the output vocabulary decreases, the ability of the trained system to generalize improves, and also its coverage.

Moreover, it reflects more accurately our intuition of what a sense is: clearly the notions of ``tree'' (with a trunk and leaves, not the mathematical graph) and ``plant'' (the living organism, not the industrial building) forms a group in our mind such that observing one sense in a context should help disambiguating the other.

\subsection{From Senses to Synsets: A First Sense Vocabulary Reduction}

WordNet is a lexical database organized in sets of synonyms called synsets. A synset is technically a group of one or more word-senses that have the same definition and consequently the same meaning. For instance, the first senses of ``eye'', ``optic'' and ``oculus'' all refer to a common synset which definition is ``the organ of sight''.

Training a WSD supervised system to predict synset tags instead of word-sense tags is a common practice \citep{yuan_2016,minh2018}, and it can be seen as a form of output vocabulary reduction based on the knowledge that is present in WordNet.

\begin{figure}[htbp]
\centering
\includegraphics[width=1.0\linewidth]{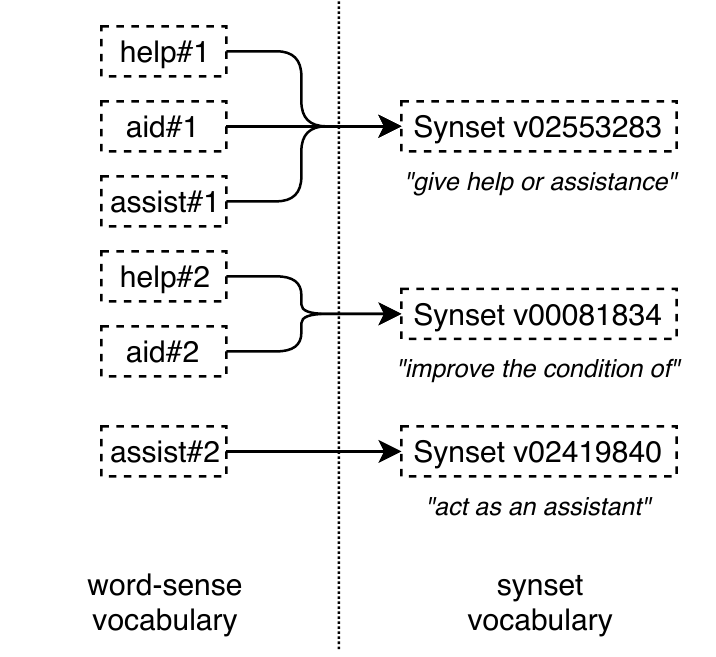}
\caption{Word-sense to synset vocabulary reduction applied on the first three senses of the words ``help'', ``aid'' and ``assist''}
\label{fig:sense_to_synset}
\end{figure}

Illustrated in \autoref{fig:sense_to_synset}, the word-sense to synset vocabulary reduction clearly helps to improve the coverage of supervised systems. Indeed, if the verb ``help'' is observed in the annotated data in its first sense, and consequently with the tag ``v02553283'', the context surrounding the target word can be used to later annotate the verb ``assist'' or ``aid'' with the same valid tag.

Once applied, the number of different labels needed to cover all senses of WordNet drops from 206\,941 to 117\,659 (approximately 43\% of reduction), and considering the SemCor, the corpus contains 26\,215 different synsets, which accounts now for 22\% of this total. The vocabulary size was reduced and the coverage improved.

Going a little further, other information from WordNet can help the system to generalize. In the next section, we describe a new method for taking advantage of the hypernymy and hyponymy relationships in order to accomplish this same idea.

\subsection{Sense Vocabulary Reduction through Hypernymy and Hyponymy Relationships}\label{sec:method}

According to \citet{polguere2003}, hypernymy and hyponymy are two semantic relationships which correspond to a particular case of sense inclusion:
the hyponym of a term is a specialization of this term, whereas its hypernym is a generalization. For instance, a ``mouse'' is a type of ``rodent'' which is in turn a type of ``animal''.

In WordNet, these relationships bind nearly every nouns\footnote{Only 2\% of the polysemous nouns of WordNet are not part of this hierarchy.} together in a tree structure that goes from the generic root, %
the node ``entity'' to the most specific leaves, for instance %
the node ``white-footed mouse''. These relationships are also present on several verbs: so for instance ``add'' is a way of ``compute'' which in turn is a way of ``reason'' which is a way of ``think''. 

For the sake of WSD, just like grouping together the senses of a same synset helps to better generalize, we hypothesize that grouping together the synsets of a same hypernymy relationship also helps in the same way. 

The general idea of our method is that the most specialized concepts in WordNet are often superfluous in order to perform WSD. 

\begin{figure}[htbp]
\centering
\includegraphics[width=0.8\linewidth]{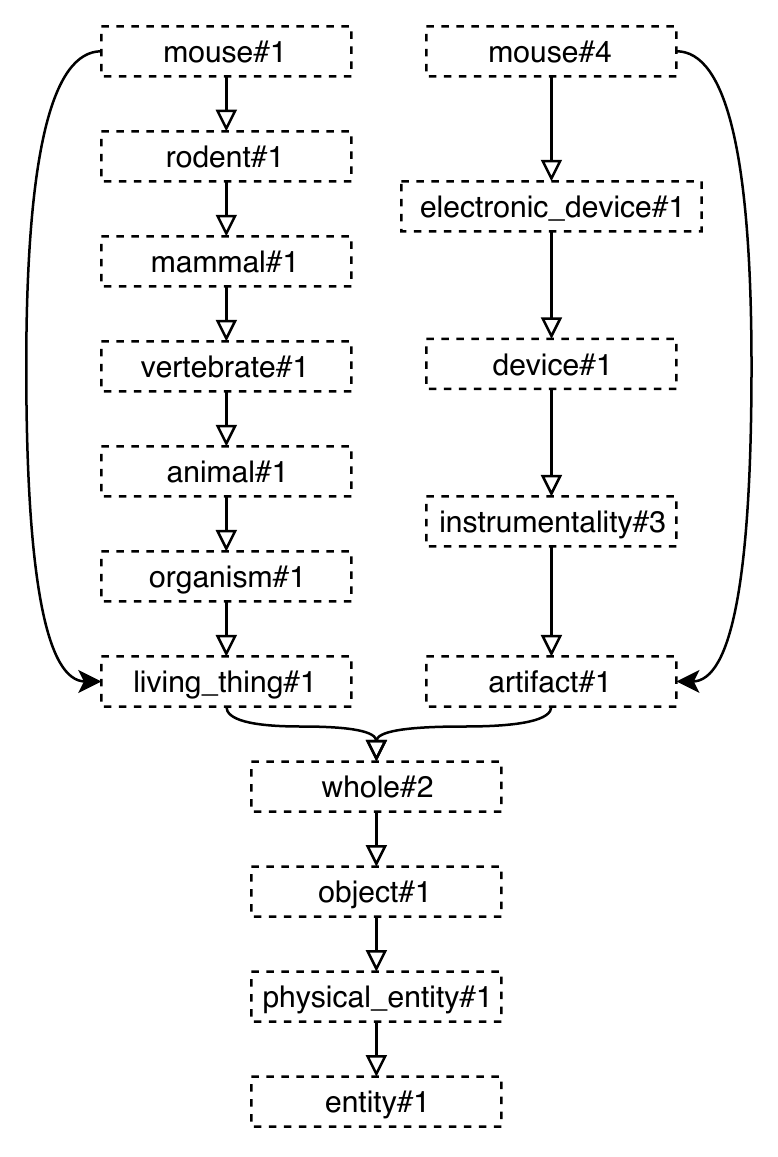}
\caption{Sense reduction trough hypernymy hierarchy applied on the first and fourth sense of the word ``mouse'' (some nodes are missing for clarity)}
\label{fig:mouse_hypernym}
\end{figure}

Indeed, consider a small subset of WordNet that only consists of the word ``mouse'', its first sense (the small rodent), its fourth sense (the electronic device), and all of their hypernyms. This is illustrated in \autoref{fig:mouse_hypernym}. We can see that every concept that is more specialized than the concepts ``artifact'' and ``living\_thing'' could be removed, and we could map every tag of ``mouse\#1'' to the tag of ``living\_thing\#1'' and we could still be able to disambiguate this word, but with a benefit: all other ``living things'' and animals in the sense annotated data could be tagged with the same sense, give examples of what is an animal and then show how to differentiate the small rodent to the hand-operated electronic device. 

In order to achieve this goal of mapping every sense to its most generic sense still allowing to differentiate the meanings of the words, we have to consider certain difficulties that are not present with the word-sense to synset vocabulary reduction.

First, contrary to the synonymy relationship which is symmetric (i.e. if A is a synonym of B then B is a synonym of A), the hypernymy relationship is not. For instance, all mice are animals, but not all animals are mice.

In addition, two different senses of a word necessarily have two different synsets, but they may have the same direct hypernym, and they generally have the same inherited hypernym at a certain point. For instance, we can distinguish the sense 1 of ``mouse'' which is a type of ``animal'' from the sense 4 which is a type of ``electronic device'', but we cannot distinguish them if we go too far into the hypernymy hierarchy, because both of them are a type of ``physical entity''.

Finally, we could think of removing a synset from the vocabulary of WordNet because it is not useful locally (from the point of view of a specific word), but it could be necessary to diferentiate the meanings of another word. %

Our method thus works in two steps:
\begin{enumerate}
    \item We mark as ``necessary'' all synsets that are the lowest nodes of the hypernymy hierarchies of the senses of all word that can still allow to discriminate the different senses of the word.
    \item We transform our sense vocabulary by mapping every synset to the lowest synset in its hypernymy hierarchy that is marked as ``necessary''. 
\end{enumerate}

The result of this method is that the most specific synsets of the tree that are not useful for discriminating are automatically removed from the vocabulary. In other words, the set of synsets that is left in the vocabulary is the smallest subset of all synsets that are necessary to distinguish every sense of every word of WordNet.

When applied on WordNet, the number of synset in the vocabulary now drops from 117\,659 to 39\,147 (approximately 66\% of reduction), and applied on the SemCor, it now contains 12\,779 different synsets, which counts for 32\% of coverage. Again, the vocabulary size has drastically decreased, and the coverage really improved. Note that if we narrow down our computation to consider only polysemic words in WordNet, the full vocabulary of all reduced synsets of WordNet is 23\,148, and the SemCor contains 9\,461 of them represented, and that is a coverage of approximately 40\%.

\section{Experiments}

In order to evaluate our vocabulary reduction method, we applied it on a classification based neural network (\autoref{sec:classif_based_wsd}) capable of classifying a word in all possible synsets of WordNet. Our architecture is very similar to \citet{raganato2017}'s BiLSTM model except for the input and output vocabulary used. Indeed, in their system, they have chosen to include their input vocabulary in their output vocabulary, so their network is able to predict both a sense tag when the target word has an entry in WordNet (nouns, verbs, adjectives and averbs), and a word tag for every other word (pronouns, articles, etc.). In our architecture, we chose to only predicts sense tags, in order to keep the output vocabulary the smallest possible.

Then we systematically trained two models:
\begin{enumerate}
    \item A baseline model that predicts a tag belonging to all the synset tags seen during training (thus using the common word-sense to synset vocabulary reduction).
    \item A second system trained under the same conditions, but with our vocabulary reduction through hypernyms algorithm applied on the training corpus.
\end{enumerate}

\subsection{Neural Architecture}

The architecture of our neural network relies on 3 layers :

\begin{enumerate}
    \item The input layer, which takes directly the words in a vector form, from a pre-trained word embeddings model. %
    \item The hidden layer, composed of bidirectional LSTM units \citep{LSTM1997}. %
    \item The output layer, which represents for each word in input, a probability distribution over all senses in the output vocabulary used, thanks to a classical softmax function.
\end{enumerate}

The cost function to minimize during the training is the cross entropy between the output layer and a one-hot vector, i.e. a vector for which all coordinates are set to $0$ except for the coordinate at the index of the target sense which is $1$. In consequence, the cost function is $-\log q[s]$, where $q[s]$ is the output of the network at the index $s$ of the target sense.

Our model always predicts a sense in output, for every input word, even for words that do not convey directly a meaning (e.g. stop-words, articles, etc.) or words that were not annotated in the training set. However, we assign a special tag \textit{<skip>} to these cases, allowing us to ignore the predictions made by the model and to not take it into account during the back-propagation step of the training. 

This behavior is the main difference between our architecture and the one introduced by \cite{raganato2017}. In their model, the gradient is computed over all words of a sentence, and those that do not have a sense in WordNet are annotated with their surface form.

In input of our network, we used the GloVe vectors \citep{pennington2014glove} pretrained on Wikipedia 2014 and Gigaword 5\footnote{\url{https://nlp.stanford.edu/projects/glove/}}. The dimension of the vectors is 300, the vocabulary size is 400\,000 and all words are lowercased. %
These vectors are also used as input in the network described by \cite{kaageback2016word}. 

For the hidden layer of recurrent units, we chose LSTM cells of size 1000 for each direction (so 2000 in total). This is approximately the same size that was used in \cite{raganato2017} (1024 per direction) and \cite{yuan_2016} (a single layer of size 2048).

Finally, we applied the regularization method Dropout \cite{Srivastava2014DSW26274352670313} between the hidden layer and the output layer, with a parameter set to $50\%$, in order to avoid overfitting during the training and to make the model more robust.

We implemented our neural network using PyTorch\footnote{\url{https://pytorch.org}}, and our code is available at the following URL: \newline
\url{https://github.com/getalp/disambiguate}

\begin{table*}[htbp]
\small
\begin{center}
\tabulinesep=2pt
\begin{tabu} to \linewidth {X[6.5lm]X[1cm]X[1cm]X[1cm]X[1cm]X[1cm]X[1cm]X[1cm]} %
\toprule
System & SE2 & SE3 & SE07 (17) & SE13 (12) & SE15 (13) & ALL & SE07 (07) \\
\midrule
SemCor, baseline & 91.15 & 96.76 & 97.58 & 91.06 & 94.78 & 93.23 & 92.84 \\
\rowcolor{gray!10} SemCor, vocabulary reduced & 98.03 & 99.19 & 99.78 & 99.15 & 98.39 & 98.75 & 98.85 \\
\midrule
SemCor+WNGT, baseline & 97.81 & 98.92 & 99.34 & 97.63 & 99.34 & 98.26 & 98.45 \\
\rowcolor{gray!10} SemCor+WNGT, vocabulary reduced & 99.74 & 99.95 & 100 & 99.76 & 99.91 & 99.83 & 99.91 \\
\bottomrule
\end{tabu}
\end{center}
\caption{Coverage of supervised systems based on the training corpus and if the vocabulary reduction algorithm is applied or not. Numbers are the percentage of words that are observed during training and hence can be annotated.}
\label{tab:coverage}
\end{table*}

\subsection{Training}

We compared our sense vocabulary reduction method on two training sets: The first is the SemCor \citep{Miller1993}, the most popular corpus that is used for training most WSD supervised systems. The second is the concatenation of the SemCor and the WordNet Gloss Tagged\footnote{\url{http://wordnetcode.princeton.edu/glosstag-files/glosstag.shtml}}. The latter is a corpus distributed as part of WordNet since its version 3.0, and it consists of all the definitions (glosses) of every synset of WordNet, with every word manually or semi-automatically sense annotated. %
We used the version of these corpora given as part of the UFSAC~2.1 resource\footnote{\url{https://github.com/getalp/UFSAC}} \citep{vialhal01718237}, a set of gathered publicly available sense annotated corpora converted into a clean and unified format.

We performed every training for 20 epochs. That is, the whole training set has been read 20 times. At the beginning of each epoch we shuffled the training set. We evaluated our model at the end of every epoch on a development set, and we kept only the one which obtained the best F1 WSD score. The development set was composed of 4\,000 random sentences taken from the WordNet Gloss Tagged for the models trained on the SemCor, and 4\,000 random sentences extracted from the training set for the other models.

We trained with mini-batches of 100 sentences, truncated to 80 words, and padded with zero vectors from the end, and we used Adam \cite{KingmaB14}, with the same default parameters described in their article as the optimization method, except for the learning rate that we set to $0.0001$ (10 times smaller than the default value).

All models have been trained on Nvidia's Titan X GPUs. The approximate training times of individual models, depending on the training corpus and if the vocabulary reduction method was applied, are displayed in the following table: 

\begin{table}[h]
\small
\centering
\tabulinesep=2pt
\begin{tabu} to \linewidth {X[1lm]|X[1cm]|X[1.3cm]}
Model & SemCor & SemCor+WNGT \\
\midrule
baseline & 50mn & 360mn (6h) \\
\midrule
vocabulary reduced & 30mn & 160mn (2h40) \\
\end{tabu}
\label{tab:times}
\end{table}

\subsection{Disambiguation}

In order to disambiguate an input sequence of words using the trained model, we followed the following steps: 

First, each word of the sequence is lowercased and transformed into a vector using the pre-trained word embeddings model, then the sequence of vectors is given as input to our model.

Then, we annotate each word with the one among its possible senses which has the maximum probability. We first map each sense to its synset in the case of the baseline model, or map each sense to its reduced synset, in the case of the sense vocabulary reduced model, according to the method described in \autoref{sec:method}, then we select the one which has the maximum value in the output layer of the model.

Finally, if no sense is assigned, because no instance of the word has been observed in the training data, a back-off is performed. We chose the most common one which is to assign the first sense in WordNet. 

\begin{table*}[htbp]
\small
\begin{center}
\tabulinesep=2pt
\begin{tabu} to \linewidth {X[5lm]X[1cm]X[1cm]X[1cm]X[1cm]X[1cm]X[1cm]X[1cm]} %
\toprule
System & SE2 & SE3 & SE07 (17) & SE13 (12) & SE15 (13) & ALL & SE07 (07)\\
\midrule
Our system (SemCor, baseline) & \tbf{73.50} ($\pm 0.26$) & \tbf{70.86} ($\pm 0.42$) & \tbf{62.47} ($\pm 1.05$) & 67.55 ($\pm 0.41$) & \tbf{71.60} ($\pm 0.52$) & 70.51 ($\pm 0.16$) & 83.77 ($\pm 0.28$)\\
\rowcolor{gray!10} Our system (SemCor, vocabulary reduced) & 72.55 ($\pm 0.50$) & 70.39 ($\pm 0.41$) & 61.46 ($\pm 1.28$) & \tbf{70.77} ($\pm 0.55$) & 71.32 ($\pm 0.49$) & \tbf{70.77} ($\pm 0.21$) & \tbf{84.48} ($\pm 0.27$) \\
\midrule
Our system (SemCor+WNGT, baseline) & \textcolor{red}{\tbf{74.40}} ($\pm 0.51$) & \tbf{70.82} ($\pm 0.49$) & 62.48 ($\pm 1.29$) & 70.82 ($\pm 0.58$) & \textcolor{red}{\tbf{74.46}} ($\pm 0.71$) & \textcolor{red}{\tbf{71.93}} ($\pm 0.35$) & 84.95 ($\pm 0.42$) \\
\rowcolor{gray!10} Our system (SemCor+WNGT, vocabulary reduced) & 74.28 ($\pm 0.40$) & 69.58 ($\pm 0.50$) & \tbf{64.46} ($\pm 1.31$) & \textcolor{red}{\tbf{71.47}} ($\pm 0.79$) & 73.71 ($\pm 0.68$) & 71.79 ($\pm 0.29$) & \textcolor{red}{\tbf{85.79}} ($\pm 0.26$) \\
\midrule
\citep{yuan_2016} (SemCor, LSTM) & 73.6 & 69.2 & 64.2 & 67.0 & 72.1 & - & 82.8 \\
\rowcolor{gray!10} \citep{yuan_2016} (SemCor, LSTM + LP) & 73.8 & \textcolor{red}{71.8} & 63.5 & 69.5 & 72.6 & - & 83.6 \\
\citep{raganato2017} (SemCor, BLSTM) & 71.4 & 68.8 & *61.8 & 65.6 & 69.2 & 68.9 & - \\
\rowcolor{gray!10} \citep{raganato2017} (SemCor, BLSTM + att. + LEX + POS) & 72.0 & 69.1 & *\textcolor{red}{64.8} & 66.9 & 71.5 & 69.9 & 83.1 \\
\citep{iacobacci2016embeddings} (IMS+emb) (Results from \citet{raganato2017}) & 72.2 & 70.4 & 62.6 & 65.9 & 71.5 & 70.1 & 81.9 \\
\midrule
First sense baseline & 65.6 & 66.0 & 54.5 & 63.8 & 67.1 & 65.5 & 78.9\\
\bottomrule
\end{tabu}
\end{center}
\caption{F1 scores (\%) obtained by our systems against the state of the art on the English WSD tasks of the evaluation campaigns SensEval 2 (SE2), SensEval 3 (SE3), SemEval 2007 (SE07) task 7 and 17, SemEval 2013 (SE13) task 12, SemEval 2015 (SE15) task 13 and the corpus composed of the concatenation of all previous ones (ALL) except SE07 task 7. Results in \tbf{bold} are the best results from using the sense vocabulary reduction or not. Results in \textcolor{red}{red} are to our knowledge the best results obtained on the task. Our results are the mean scores of 20 individual systems, with the standard deviation given in parenthesis. Results prefixed by a star (*) was obtained on the development corpus used during the training.}
\label{tab:scores}
\end{table*}

\begin{table*}[htbp]
\small
\begin{center}
\tabulinesep=2pt
\begin{tabu} to \linewidth {X[5lm]X[1cm]X[1cm]X[1cm]X[1cm]X[1cm]X[1cm]X[1cm]} %
\toprule
System (ensemble) & SE2 & SE3 & SE07 (17) & SE13 (12) & SE15 (13) & ALL & SE07 (07) \\
\midrule
Train on SemCor, baseline & 73.93 & 70.81 & 63.74 & 67.27 & 72.27 & 70.77 & 83.77 \\
\rowcolor{gray!10} Train on SemCor, vocabulary reduced & 73.05 & 70.59 & 61.32 & 71.23 & 71.60 & 71.18 & 84.65 \\
Train on SemCor+WNGT, baseline & 74.72 & \tbf{71.08} & 62.86 & 71.11 & 74.83 & 72.23 & 85.01 \\
\rowcolor{gray!10} Train on SemCor+WNGT, vocabulary reduced & \textcolor{red}{\tbf{75.15}} & 70.11 & \textcolor{red}{\tbf{66.81}} & \textcolor{red}{\tbf{72.63}} & \textcolor{red}{\tbf{74.46}} & \textcolor{red}{\tbf{72.74}} & \textcolor{red}{\tbf{86.02}} \\
\bottomrule
\end{tabu}
\end{center}
\caption{F1 scores (\%) obtained by our system with an ensemble of 20 models trained separately. Results in \tbf{bold} are the best results from all our systems. Results in \textcolor{red}{red} are to our knowledge the best results obtained on the task.}
\label{tab:scores2}
\end{table*}

\subsection{Evaluation}

We evaluated our models on all evaluation corpora commonly used in WSD, that is the WSD tasks of the evaluation campaigns SensEval/SemEval. 
We used the fine-grained evaluation corpora from the evaluation framework of \citet{raganatocamachocolladosnavigli2017}, which consists of SensEval~2 \citep{Edmonds2001}, SensEval~3 \citep{W040811}, SemEval~2007 task~17 \citep{Pradhan2007}, SemEval~2013 task~12 \citep{Navigli2013} and SemEval~2015 task~13 \citep{moronavigli2015}, as well as their ``ALL'' corpus consisting of the concatenation of all previous ones. We also compared our result on the coarse-grained task~7 of SemEval~2007 \citep{Navigli2007} which is not present in the evaluation framework. We used the version of these corpora from the UFSAC~2.1 resource\footnote{\url{https://github.com/getalp/UFSAC}}, the sense inventory used for the sense annotations is WordNet~3.0.

For each evaluation, we trained 20 separated models, and we give two scores: First, the mean of the F1 scores obtained by the models, along with its standard deviation. Then, the score obtained by an ensemble of the models. For the ensemble, we averaged the predictions of all individual models through a geometric mean, a common practice that is used for instance in machine translation \citep{Sutskever2014,gehring2017}.

\subsection{Results}

The scores obtained by our systems using a single trained model compared to the state-of-the-art systems \cite{yuan_2016,raganato2017,iacobacci2016embeddings}, along with the first sense baseline are present in table~\ref{tab:scores}. The scores obtained by our ensemble of models are given in \autoref{tab:scores2}.

In \autoref{sec:method}, we showed that our vocabulary reduction method improves the coverage of supervised systems over all WordNet vocabulary. In \autoref{tab:coverage}, we can see that this coverage improvement holds true on the evaluation tasks, for both training sets. On the total of 7\,253 words to annotate for the corpus ``ALL'', the baseline system trained on the SemCor only is incapable of annotating 491 of them, and with the vocabulary reduction applied this number drops to 91. When adding the WordNet Gloss Tagged to the training set, this number is 126 for the baseline system, and with the vocabulary reduction, only 12 words cannot be annotated. 

Now if we look at the results in \autoref{tab:scores}, the difference of scores obtained by our system using the sense vocabulary reduction or not is overall not significant (regarding the ``ALL'' column). However we can notice a very large gap on the SemEval 2013 task, especially when the SemCor is used alone for training. This can be explained by the fact that this corpus is only composed of nouns, and our method for vocabulary reduction targets this part of speech principally. This is also the task where the coverage was improved the most by our method, as it can be seen in \autoref{tab:coverage}.   

In comparison with the other works, our systems trained on the SemCor alone expose results comparable with the best system of \citet{yuan_2016}, which is trained on the same corpus and augmented with a semi-supervised method. When we add the WordNet Gloss Tagged to the training data however, we obtain systematically state of the art results on all tasks except on SensEval~3. Once again, the sense reduction method does not consistently improves or decreases the score on every task, and in overall (task ``ALL''), the result is roughly the same as without sense reduction applied.

Finally, in \autoref{tab:scores2} we show the results of our system ensembling 20 models by averaging the output of their last layer. As we can see, ensembling is a very efficient method in WSD as it improves systematically all our results. Interestingly, with ensembles, the scores are significantly higher when applying the vocabulary reduction algorithm. One possible interpretation is that individual models might be more frequently ``lost'' in the sense that with the sense vocabulary reduction applied, a lot of words are annotated with the same tag, and it can make the trained model ``unsure'' about the decisions it make. Ensemble of models tends to prevent this problem by favoring the most probable decisions of the models. 

\section{Conclusion}

In this paper, we presented a new method that improves the coverage and the capacity of generalization of most supervised systems, by narrowing down the number of different sense in WordNet in order to keep only the senses that are essential for differentiating the meaning of all words present in the lexical database.

By considering that a same sense tag may be applied to many different words, this method also captures in a intuitive way a better representation of what is a sense for the task of word sense disambiguation.

We implemented a state of the art neural network for WSD and we showed that this method really improves the overall results using the same training corpus, especially when making ensembles of models, and also especially when disambiguating nouns. 

We trained two sets of systems, one relying on the SemCor alone, and one with the addition of the WordNet Gloss Tagged corpus, and in this last configuration we obtained our best results that significantly outperform the state of the art on most WSD tasks.

With the combination of our sense vocabulary reduction method through the hypernymy hierarchy of WordNet and the addition of the WordNet Gloss Tagged to the set of training corpora, the coverage of our supervised system is almost 100\% on most WSD tasks, and so this provides a solid alternative to the automatic or semi-automatic creation of sense annotated corpora, and this nearly eliminates the need for a first sense backoff in WSD supervised systems. 

\clearpage

\bibliographystyle{acl_natbib}

\bibliography{biblio}

\end{document}